\documentclass[double]{article}
\usepackage{times}
\usepackage{algorithm}
\usepackage{algorithmic}
\usepackage{wrapfig}
\usepackage{verbatim}
\usepackage{amsmath}
\usepackage{graphicx}
\usepackage{subcaption}
\usepackage[active]{srcltx}
\usepackage{color}
\usepackage{lineno}
\usepackage{dirtytalk}
\usepackage{amsthm}
\usepackage{authblk}
\usepackage{listings}
\usepackage{tikz}
\usepackage{longtable}
\usetikzlibrary{shapes.geometric, arrows}

\usepackage{epsfig,graphicx,amsfonts}
\usepackage{amsmath,amsthm,amssymb}
\usepackage{xcolor}

\usepackage{adjustbox}
\usepackage{multirow}
\usepackage{setspace}
\usepackage{hyperref}
\usepackage{comment}

\long\def\comment #1\commentend{}

\begin{document}

\title{\Large Pulling the Carpet Below the Learner's Feet: Genetic Algorithm To Learn Ensemble Machine Learning Model During Concept Drift}

\author{Teddy Lazebnik$^{1,2,*}$\\
\(^1\) Department of Mathematics, Ariel University, Ariel, Israel\\
\(^2\) Department of Cancer Biology, Cancer Institute, University College London, London, UK\\
\(^*\) Corresponding author: lazebnik.teddy@gmail.com \\
}

\date{ }

\maketitle 

\begin{abstract}
\noindent
Data-driven models, in general, and machine learning (ML) models, in particular, have gained popularity over recent years with an increased usage of such models across the scientific and engineering domains. When using ML models in realistic and dynamic environments, users need to often handle the challenge of concept drift (CD). In this study, we explore the application of genetic algorithms (GAs) to address the challenges posed by CD in such settings. We propose a novel two-level ensemble ML model, which combines a global ML model with a CD detector, operating as an aggregator for a population of ML pipeline models, each one with an adjusted CD detector by itself responsible for re-training its ML model. In addition, we show one can further improve the proposed model by utilizing off-the-shelf automatic ML methods. Through extensive synthetic dataset analysis, we show that the proposed model outperforms a single ML pipeline with a CD algorithm, particularly in scenarios with unknown CD characteristics. Overall, this study highlights the potential of ensemble ML and CD models obtained through a heuristic and adaptive optimization process such as the GA one to handle complex CD events. \\ 

\noindent
\textbf{Keywords}: automatic machine learning; heuristic optimization; concept drift; ensemble machine learning.
\end{abstract}

\maketitle \thispagestyle{empty}
\pagestyle{myheadings} \markboth{Draft:  \today}{Draft:  \today}
\setcounter{page}{1}

\section{Introduction}
\label{sec:introduction}
Data-driven models, in general, and machine learning (ML) models, in particular, have gained popularity over recent years with increased usage of such models across the scientific and engineering domains \cite{teddy_ml_1,teddy_ml_2,teddy_ml_3,teddy_ml_4,teddy_ml_5,teddy_ml_6}. While ML models show promising results in the lab as well as in realistic scenarios, deployed ML models experience a wide range of challenges in production settings. A common and important challenge these models encounter is adapting to dynamic and evolving environments \cite{cd_1,cd_2}. In particular, concept drift (CD), the phenomenon wherein the statistical properties of the target variable change over time, poses a significant hurdle to the stability and performance of learning-based models \cite{cd}. The dynamic nature of real-world data introduces uncertainties, necessitating the continuous adaptation of models to maintain their relevance and accuracy. Recent research in the CD domain focuses on addressing three main challenges. Precisely identify CD within unstructured and noisy datasets \cite{cd_1_1,cd_1_2,cd_1_3}, to comprehensively comprehend CD in a quantifiable and interpretable manner \cite{cd_2_1,cd_2_2}, and to respond effectively to CD \cite{is_2,cd_3_1}. 

A genetic algorithm (GA) is a search and optimization technique inspired by the principles of natural selection and genetic inheritance \cite{ga_intro,intro_genetic_2}. It operates by iteratively evolving a population of potential \say{solutions} to a problem through mechanisms such as selection, crossover, and mutation, mimicking the process of biological evolution to find optimal or near-optimal solutions \cite{ga_engineering,ga_medicine}. The motivation behind employing genetic algorithms lies in their ability to efficiently explore large solution spaces, enabling the discovery of diverse and effective solutions that may be elusive through traditional optimization methods \cite{ga_intro_1,ga_intro_2,ga_intro_3}. In particular, GAs have been adapted to the realm of ML to find a well-performing ML pipeline \cite{tpot}, solve a symbolic regression task \cite{ga_sr_1,ga_sr_2,ga_sr_3}, or even as part of ML ensemble-based model \cite{ga_ml_1,ga_ml_2}. 

Generally speaking, GAs hold significance in ML for three primary reasons. First, they operate in discrete spaces, making them applicable in scenarios where gradient-based methods are impractical \cite{ga_r_1}. Second, GAs function as reinforcement learning algorithms, evaluating the performance of a learning system based on a singular metric, commonly referred to as the \say{fitness} function, in contrast to approaches like back-propagation where different parts of the model received different optimization signals. This characteristic makes them particularly useful in situations where performance measurement is the sole available information \cite{ga_r_2_1,ga_r_2_2}. Third, GAs involve a population, making them suitable for scenarios where the desired outcome is not a single model but a set of models, as exemplified in learning within multi-agent systems \cite{ga_r_3}. 

To this end, GAs can be used to overcome CD \cite{ga_cd_1,ga_cd_2}. In this study, we focused on the last point as in the context of deployed ML-based solution where it is trained on initial data and more (tagged) data is gathered over time, GAs can be used to respond to changes in dynamics. Intuitively, one can think of CD as the change between the source (\(x\)) and target (\(y\)) features where some model (\(f\)) is applied at two points in time (\(t_1, t_2\)), such that \(||f_{t_1}(x) - y|| \leq ||f_{t_2}(x) - y ||\). Hence, as new data is introduced to the learner, a mechanism should alter the model \(f\) aiming to obtain  \(||f^1_{t_1}(x) - y|| = ||f^2_{t_2}(x) - y ||\) where \(f^2\) is originated from \(f^1\) and altered to achieve the above condition. Following this line of thought, we propose a two-level ML model obtained using a GA algorithm which has a global ML model with a CD detector that operates as an ensemble model for a population of ML models, each governing subset of the data and self-adoptive according to their own CD detector models. 

The rest of the paper is organized as follows. Section \ref{sec:rw} presents an overview of CD properties, challenges, and previous solutions as well as the recent developments in the field of GAs and Ensemble ML. Sectio \ref{sec:background} provides a technical background later used in the model definition. Section \ref{sec:model} formally outlines the task definition, the proposed algorithm based on GA, and its applicative improvement in the form of using automatic ML models with a divide-and-conquer approach. Section \ref{sec:experimental_setup} introduces the experimental setup used to explore the proposed algorithm. Section \ref{sec:results} shows the obtained results. Finally, Section \ref{sec:discussion} discusses the results with their potential applicative usage as well as the limitations of this study and possible future work.

\section{Related Work}
\label{sec:rw}
In this section, we present approaches in the field of CD adoption followed by an overview of the GA models used in dynamical systems. Afterward, we review several cases where GA is used in the context of CD and ensemble ML models.   

\subsection{Concept drift}
A discussion about CD contains two interconnected dichotomy aspects - the theoretical and applied aspects of defining, detecting, and tacking CD. It is more often than not the applied aspect that governs the broad interest in CD as ML users experience first-hand the challenges that come with CD in their respective tasks \cite{cd_sec_1}. Indeed, in a dynamic world, nothing is constant. For instance, let us consider a supply chain distribution system responsible for distributing a company's products between its physical stores. Would one should expect that a model that was trained before COVID-19 \cite{first_teddy_paper}, would work equally well during or even after the COVID-19 pandemic? It is reasonably easy to assume that due to these kinds of unforeseen circumstances, user behavior would change a lot, as indeed happened in practice \cite{cd_example_1,cd_example_2}. 

As such, a field of repetitive ML model adoption has been proposed where new data is used to re-train ML models to detect, capture, and utilize the changes in the data over time, practically addressing CD \cite{ml_retrain_2,ml_retrain_1}. One of the most direct approaches to address CD involves retraining a new model with the latest data to replace the outdated model and dynamics constructed it \cite{ml_window_1,ml_window_2}. This method necessitates an explicit CD detector to determine when model retraining is required. This approach does not work well when the change over time is relatively slow and \say{smooth} and excels in hard shifts in the system's dynamics. The complementary approach is to use a \say{window} strategy where the model is retrained on the latest data with some fixed size. A more sophisticated example of this approach is employed by \textit{Paired Learners} which utilizes two learners - the stable learner and the reactive learner \cite{paired_learners}. If the stable learner consistently misclassifies instances correctly identified by the reactive learner, signaling a new concept, the stable learner is replaced with the reactive learner. 

These two approaches cover the basic ideas of CD overcoming in ML. That said, each approach raises a new computational challenge one needs to tackle. The first approach requires the accurate detection of CD while the latter challenges the user to find and use the optimal window size. On top of that, as each approach is appropriate to different types of CD, choosing the appropriate one for each case is a challenge in itself. This fertile soil was the base of multiple solutions. 

Initially, attempts to find the optimal window size have been conducted as a compromise must be made in determining the suitable window size. A smaller window effectively mirrors the most recent data distribution, while a larger window affords more data for training a new model. Consequently, \cite{ml_window_sol} proposed ADWIN, an algorithm that dynamically adjusts subwindow sizes based on the rate of change between sub-windows, eliminating the necessity for users to predefine a fixed window size. After determining the optimal window cut, the window containing outdated data is discarded, facilitating the training of a new model with the latest window data.

Moving beyond mere model retraining, researchers have delved into the integration of the drift detection process with the retraining mechanism tailored for specific ML algorithms rather than an \say{one method to rule them all} approach. For example, \cite{delm} proposed DELM which extends the conventional ELM algorithm to handle concept drift by adaptively modifying the number of hidden layer nodes. Moreover, instance-based lazy learners also show promising results for CD handling \cite{lazy_cd}. For example, \cite{cd_example_last} proposed NEFCS, a kNN-based adaptive model, that utilizes a competence model-based drift detection algorithm to identify drift instances in the case base and distinguish them from noise instances. 

\subsection{Genetic algorithm}
GAs belong to a category of approaches commonly known as evolutionary computation methods, employed in adaptive aspects of computation such as search, optimization, machine learning, and parameter adjustment \cite{ga_rw_intro}. What distinguishes these approaches is their characteristic reliance on a population of potential solutions. Unlike most search algorithms that focus on modifying a single candidate solution to enhance its performance, evolutionary algorithms dynamically adapt entire populations of candidate solutions to address the problem at hand. Drawing inspiration from biological populations, these algorithms incorporate selection operators to amplify the number of superior solutions within the population while diminishing the presence of inferior ones \cite{selection_operator,feasibility_operator}. Additionally, they employ other operators to generate novel solutions. The variability among these algorithms lies in the standard representation of problems and the nature and relative significance of the operations introducing new solutions \cite{cross_over,sac_crossover,ring_crossover}.

GAs have found application across diverse domains, including engineering \cite{ga_engineering}, medicine \cite{ga_medicine}, and economics \cite{genetic_cost_optimization}. For instance,  \cite{feasibility_operator} addressed the challenge of optimizing sequences of machines and their corresponding operations for process planning optimization. Employing GAs, the authors derived feasible processes initially and subsequently identified the optimal process from this set of viable alternatives. Similarly, \cite{selection_operator} investigated and assessed the utilization of GAs under various constraints in process route sequencing and astringency. The authors revamped the GA, encompassing the development of coding strategies, the evaluation operator, and the fitness function. Their findings demonstrated that these modified GAs could effectively fulfill the requirements of sequencing tasks and meet the criteria for astringency. In another study, \cite{genetic_cost_optimization} introduced an optimization scheme based on GAs to enhance Atkinson fuel engine models, specifically targeting fuel consumption reduction \cite{genetic_cost_optimization,atkinson_fuel_engines}. GAs were chosen due to the high-dimensionality and non-linearity of the optimization system, rendering classical methods time and resource-intensive. Furthermore, the authors proposed GAs as a financial model, illustrating their applicability in learning signal utilization, making inferences from market-clearing prices, and assessing the worthiness of acquiring a signal \cite{genetic_decision_making}. In the economic domain, \cite{teddy_economy_ga} presented an agent-based model with a heterogeneous population and genetic algorithm-based decision-making to model and simulate an economy with taxation policy dynamics. Furthermore, for the clinical domain, GAs have been widely used as well. For example, \cite{teddy_cells} used GA to obtain the parameters of a partial differential equations-based model describing an immunotherapy treatment for bladder cancer.  

\subsection{Genetic algorithm for concept drift}
A growing body of work finds the usage of GA to detect and adapt to CD promising in a wide range of tasks and data settings \cite{cd_ts_review}. In particular, in realistic applications, CD is of interest when new data is obtained once an initial ML model is obtained \cite{cd_ts_1,cd_ts_2}. 

\cite{eacd} introduces an innovative ensemble learning approach relying on evolutionary algorithms to address diverse forms of concept drifts in non-stationary data stream classification tasks. The authors employ random feature subspaces drawn from a feature pool to construct distinct classification types within the ensemble. Each type comprises a finite set of classifiers (decision trees) constructed at various instances throughout the data stream. Utilizing an evolutionary algorithm, specifically replicator dynamics, the system adapts to varying concept drifts by enabling types with superior performance to expand and those with inferior performance to diminish in size.

\cite{ga_cd_rw_1} proposed a novel Density-based method for Clustering Data streams employing Genetic Algorithm (DCDGA). This approach leverages a GA to optimize parameters, specifically the cluster radius and minimum density threshold, ensuring more precise coverage of density clusters. Additionally, a Chebychev distance function is introduced to compute the distance between the center of Core Micro-Clusters (CMCs) and the incoming data points. The authors evaluated DCDGA on both artificial and real datasets and showed that the experimental results were compared with another online density-based clustering in the field.

\cite{ga_cd_rw_2} introduces a predictive model for temporal data with a numerical target, utilizing GA to capture changes in a dataset caused by concept drift. In the presence of environmental changes, which stands for the CD in these settings, the author's proposed algorithm responds by clustering the data and subsequently creating nonlinear models that characterize the formed clusters. These nonlinear models serve as terminal nodes within the GA model trees.

\cite{ga_cd_rw_3} developed a spam detection system that examines the evolution of features. The author's proposed method encompasses three key steps. First, training a spam classification model; second, detecting CD using a new strategy that analyzes feature evolution based on the similarity between feature vectors extracted from training and test data; and finally, knowledge transfer learning. In the last step, the focus is on determining what knowledge to transfer, how to transfer it, and when to execute the knowledge transfer process.

\subsection{Ensemble machine learning models}
Ensemble ML methods leverage multiple ML algorithms to generate weak predictive results by extracting features through diverse projections on data. These results are then fused using various voting mechanisms to achieve superior performance compared to that obtained from any individual algorithm in isolation \cite{eml_1}. Indeed, ensemble ML models show superior results on a wide range of tasks \cite{eml_2}. The fundamental concept of a standard ensemble ML model involves two stages: producing prediction outcomes through numerous weak classifiers and consolidating these multiple results into a consistency function to obtain the ultimate result using voting schemes. The voting scheme can range in complexity, from a simple average or majority vote for regression and classification tasks, respectively, such as for the case of the Random Forest model \cite{rf_ml} which is based on a set (forest) of Decision Tree models \cite{dt_ml} to being an ML or deep learning model by itself \cite{eml_3}.

The weak prediction ML models in the set of an ensemble method differ from one another by one or more of the following three properties. First, the samples provided to the model, the features provided to the model, and even the ML model itself \cite{eml_4}. These changes allow each weak prediction model to focus on a less complex pattern in the data and excel in capturing it. Hence, the more weak models an ensemble model includes, the more complex patterns it can capture. However, it also increases the bias given the data training data \cite{eml_4}. Importantly, this scheme can be generalized where weak models in an ensemble model can be ensemble models by themselves. For example, imagine a Random Forest model in which each Decision Tree is replaced with a Random Forest as well. This example will result in three levels of models. 

\section{Technical Background}
\label{sec:background}
In this section, the necessary technical background, later used by the proposed model, is presented. Initially, a formal definition of CD with its two main types is provided. Next, an introduction to automatic ML (AutoML) models is presented. 

\subsection{Concept drift definition}
A learning algorithm, \(A\), observing samples with a stationary distribution would observe the training cohort in the form \((x_i , y_i)\)  such that \(x_i\) is the feature vector and \(y_i\) is the target feature. A class prediction at a specific point in time \(t^*\) would be given as \(y_{t^*}\) based on the feature vector \(x_{t^*}\). Opposed to this, a data stream may produce samples with a non-stationary distribution. In such a scenario, the \((x_i, y_i)\) is obtained by a distribution that explicitly depends on time or previous samples measured from the distribution.  Formally, a CD between two points in time \(t_0\) and \(t_1\) can be defined as: \(p_{t_0}(x_i, y_i) \neq p_{t_1}(x_i, y_i)\), where \(p_t\) is the joint distribution at time t between the feature vector \(x_i\) and the target feature \(y_i\) \cite{cd_2}. Following this definition, CD can occur due to three main reasons: the distribution of samples in the target feature can change; the distribution of samples in the target feature can change concerning the samples of the source features; and the source features distribution can change while the target feature does not. 

\begin{figure}[!ht]
    \centering
    \includegraphics[width=0.99\textwidth]{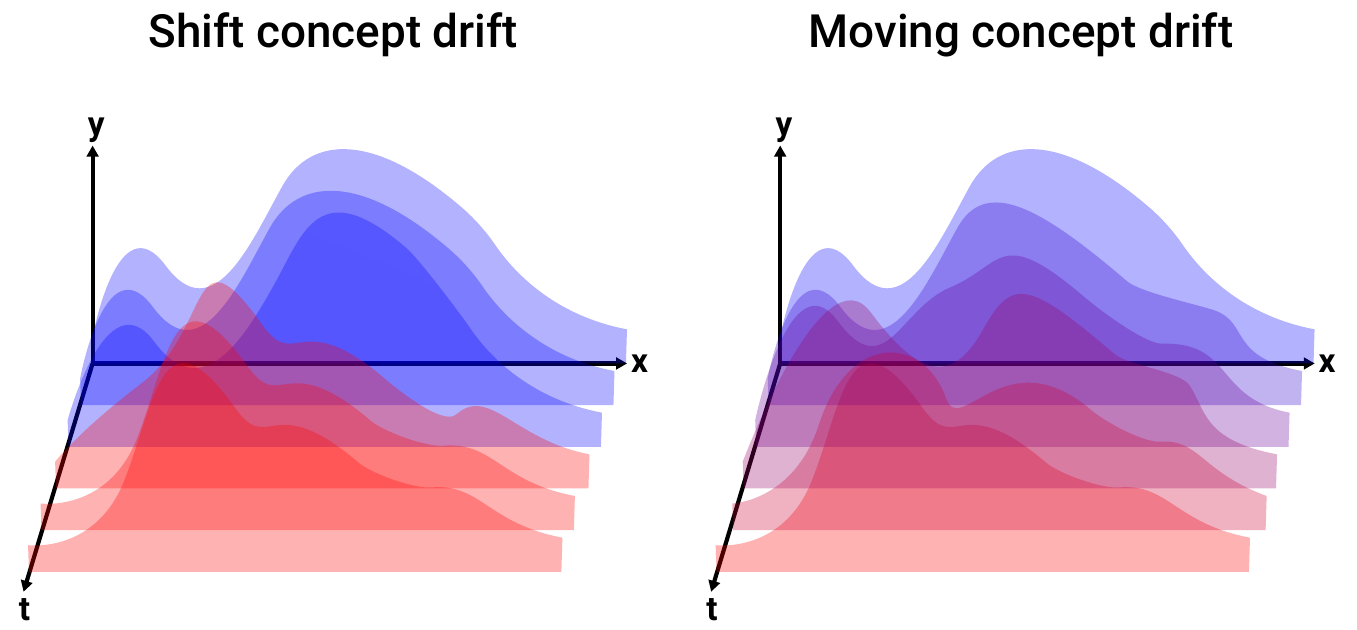}
    \caption{A schematic view of shift and moving CD. One can notice that the shift CD moves from one two-dimensional distribution (\(x,y\)) to another distribution drastically. On the other hand, the moving CD gradually alters from the same source distribution to the other distribution.}
    \label{fig:cd_types}
\end{figure}

In addition to the fact that CD occurred, the rate at which it happens is also of interest. Simply put, the rate at which CD takes place can be roughly divided into two main forms: shift and moving CD. The shift drift is associated with sudden changes in the distribution while the moving drift occurs at a much slower rate usually with multiple phases in between \cite{cd_types}. Formally, let us assume two distributions \(p_{t^*}(x_i, y_i)\) and \(p_{t^* + \Delta t}(x_i, y_i)\) of the data associated with two points in time \(t^*\) and \(t^* + \Delta t\), respectively, such that \(\Delta t \in \mathbb{R}^+\) is the time passed since the initial point in time, \(t^*\). In addition, let us assume a threshold value \(\psi \in \mathbb{R}^+\). A drifting rate is defined to be
\begin{equation}
(1 - KS(p_{t^*}, p_{t^* + \Delta t}))/ \Delta t,
\label{eq:drifting_rate}
\end{equation}
such that \(KS(a, b)\) is the p-value of a Kolmogorov-Smirnov test \cite{ks} between the two distribution \(a\) and \(b\). Fig. \ref{fig:cd_types} presents a schematic view of shift and moving CD where the shift CD moves from one two-dimensional distribution (\(x,y\)) to another drastically while the moving CD gradually alters from the same source distribution to the other distribution.

\subsection{Automatic Machine Learning}
The process of ML model development is time-consuming, requires substantial expertise, and is susceptible to human errors \cite{teddy_automl}. To this end, AutoML has emerged as a promising approach that automates many steps in the ML development process, including data pre-processing, feature engineering, model selection, and hyperparameter tuning, thereby mitigating the challenges associated with using ML models \cite{rw_automl_1,nisioti,pinto,ludwig}.

Multiple models have been proposed in recent years for Automatic machine learning \cite{fspl}. For instance, the Tree-based Pipeline Optimization Tool (TPOT) library utilizes a GA search process to identify ML pipelines based on the popular Scikit-learn library \cite{scikit_learn}. TPOT uses a tree-based representation to evolve and optimize these pipelines based on their performance, aiming to find the most effective combination for a given dataset. The library represents machine learning pipelines as tree structures, providing a flexible and hierarchical way to organize and evolve complex combinations of data preprocessing and modeling steps. The AutoSklearn library \cite{auto_sklearn,auto_sklearn_2} employs various search methods to construct an ML pipeline, also based on the Scikit-learn library. It employs meta-learning and Bayesian optimization techniques to efficiently search through various preprocessing steps, feature engineering methods, and model configurations. AutoSklearn incorporates meta-learning, leveraging information from previous ML tasks to guide the search for effective pipelines. The AutoGluon library \cite{autogluon} strategy is based on the idea of ensembling multiple models and stacking them in multiple layers. AutoGluon uses a fixed defaults (set adaptively) strategy for the search process of ML models in each layer and then combines multiple layers using the stacking and repeated bagging methods. The PyCaret library \cite{pycaret} is a Python wrapper around several popular ML libraries and frameworks which uses a multi-metric comparison of these models, in a brute-force manner, to find the best model for a given dataset and task.

\section{Ensemble Machine Learning Model For Concept Drift Data}
\label{sec:model}
\begin{figure}[!ht]
    \centering
    \includegraphics[width=0.99\textwidth]{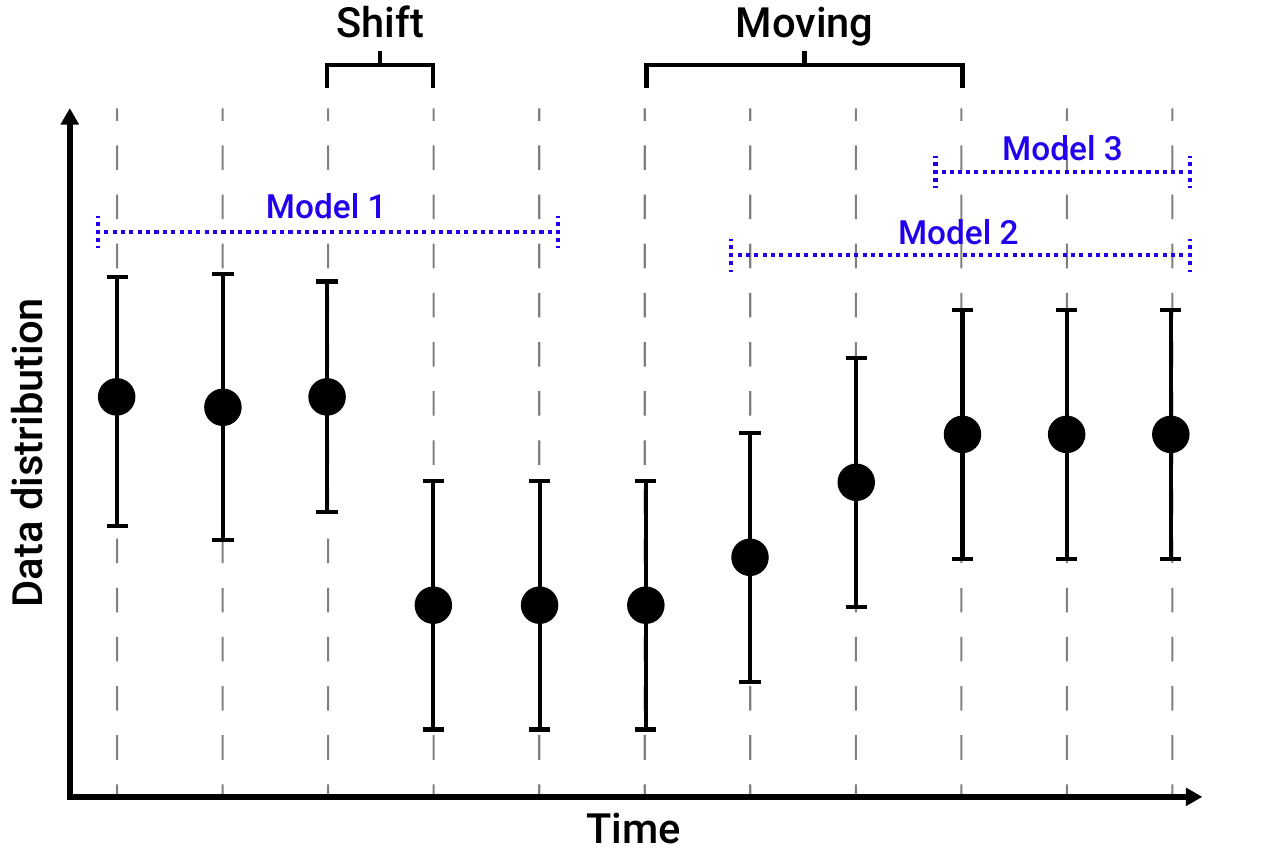}
    \caption{A schematic view of the learning problem during different CD types and possible remedy with ensemble ML model obtained using an initial search process. The distributions over time are shown as mean \(\pm\) standard deviation of some random variable, as reflected on the y-axis. The x-axis indicates steps in time. In this example, a shift CD has occurred between the third and fourth steps in time. In addition, a moving CD has occurred between the sixth and tenth steps in time. A possible ensemble model to tackle this condition would detect the shift and moving CDs and use three models, one to capture the original data between the CDs, a second model that takes into account the recent, seemingly stable, data with some "tail" of the moving CD, and a third model that based only on the recent time without CD. }
    \label{fig:idea}
\end{figure}

In this section, we outline the proposed GA-based solution for CD in data streams based on ensemble machine learning models. The proposed model is based on the fact that there are existing feasible, and even well-performing, solutions for CD types when these occur individually or under some assumptions. Furthermore, as there is no one solution to rule out all CD types, one is required to find an appropriate solution for each case. However, using multiple models, it is possible to activate one or a subset of these models as the dynamics of the system alter over time. Fig. \ref{fig:idea} presents a schematic view of the learning problem during different CD types and possible remedy with an ensemble ML model obtained using an initial search process. Intuitively, one can perform a two-step optimization process where the first step is responsible for the ensemble configuration in terms of the model and the data it obtained during the training phase, and the second step is to find and train an ML model with the CD-handling method that best suits the data it obtained.    

\subsection{Task definition}
Initially, to measure how well a population of ML models handles a CD scenario, a metric needs to be defined. Intuitively, the population of ML models should perform well on the initial dataset at some time, \(t\), and not lose this performance over some fixed duration of time. Hence, let us consider a population of ML models, \(\textbf{M} := [M_1, \dots, M_k]\) and a dataset that increases over time \(D(t) \in \mathbb{R}^{n \times m}\), such that each model (\(M_i \in M\)) obtains a subset, \(D_i \subset D(t)\) at some point in time \(t\). For a given event horizon \(\tau \in \mathbb{R}^+\) and a performance metric, \(\psi\), the CD handling performance of the ML models population, \(L(M, D(t))_{\psi, \tau}\), is defined as follows:
\begin{equation}
    L(M, D(t))_{\psi, h} := \omega_1 \psi(M, D(t)) - \omega_2 \big ( \psi(M, D(t)) - \psi(M, D(t+\tau)) \big ),
    \label{eq:metric}
\end{equation}
where \(\omega_1 \in \mathbb{R}^+\) and \(\omega_2 \in \mathbb{R}^+\) are the weights of the model's performance at the end of the training phase and the weight of the CD's influence on the models' performance, respectively. Based on this definition, one can formalize an optimization task to find the population of ML models as follows:
\begin{equation}
    \underset{M}{\max} \;  L(M, D(t))_{\psi, \tau}.
    \label{eq:optimization_task}
\end{equation}

\subsection{Genetic algorithm based solution}
One can solve Eq. (\ref{eq:optimization_task}) in multiple ways. Naively, assuming a finite number of ML models and a maximum number of models in the population, one can theoretically brute force the optimization task. Nonetheless, due to the extremely large set of possible solutions, such an approach is infeasible in practice \cite{brute_force_auto_ml_bad_idea}. In an opposite manner, one can try this optimization task analytically, however, without further assumptions over either \(M\), \(D(t)\), or \(\psi\) it seems infeasible to obtain such a solution. Thus, one can use a heuristic approach to solve Eq. (\ref{eq:optimization_task}). In this study, we suggest an adoption of the classical GA for this task. 

Formally, we assume that the population of ML models, \(M\), is defined by both the number of models, \(n\), as well as the models themselves. Each ML model, \(M_i \in M\), is contracted from a feature engineering algorithm, a supervised ML algorithm, and a hyperparameters tuning algorithm, each of these algorithms is chosen from a pre-defined and finite set of algorithms. In addition, each ML pipeline model is affected by the data it is provided with during the training and testing phases. As such, each model should be provided with a non-empty subset, \(d_i \subset D(t)\) that is used to train the ML pipeline model. Based on this representation, solving for Eq. (\ref{eq:optimization_task}) would optimize \(\psi(M, D(t))\) while is not considered \(\psi(M, D(t)) - \psi(M, D(t+\tau))\) as no remedy for the treating change over time for \(D(t)\) is considered. Consequently, one should include a CD detection algorithm for each ML pipeline model as well as for the entire population. As such, \(M\) can be represented by \(M := \{(f_g, m_g, h_g, D(t), cd_g), n, \big ((f_1, m_1, h_1, d_1, cd_1), \dots, (f_{n-1}, m_{n-1}, h_{n-1}, d_{n-1}, cd_{n-1}), (f_n, m_n, h_n, d_n, cd_n) \big )\) such that \(f_i \in \mathbb{F}\) is the feature engineering algorithm, \(m_i \in \mathbb{M}\) is the supervised ML algorithm, \(h_i \in \mathbb{H}\) is the hyperparameters tuning algorithm, \(d_i \subset D(t)\) is subset of the data used to train \((f_i, m_i, h_i)\), \(cd_i \in \mathbb{CD}\) is the CD detection algorithm. In addition, \((f_g, m_g, h_g, cd_g)\) is the global ML pipeline model responsible which gets as input the output of \((f_i, m_i, h_i)\) and \(cd_i\) for each \(i \in [1, \dots, n]\) and making the final prediction of the \(M\) model. 

\begin{figure}[!ht]
    \centering
    \includegraphics[width=0.99\textwidth]{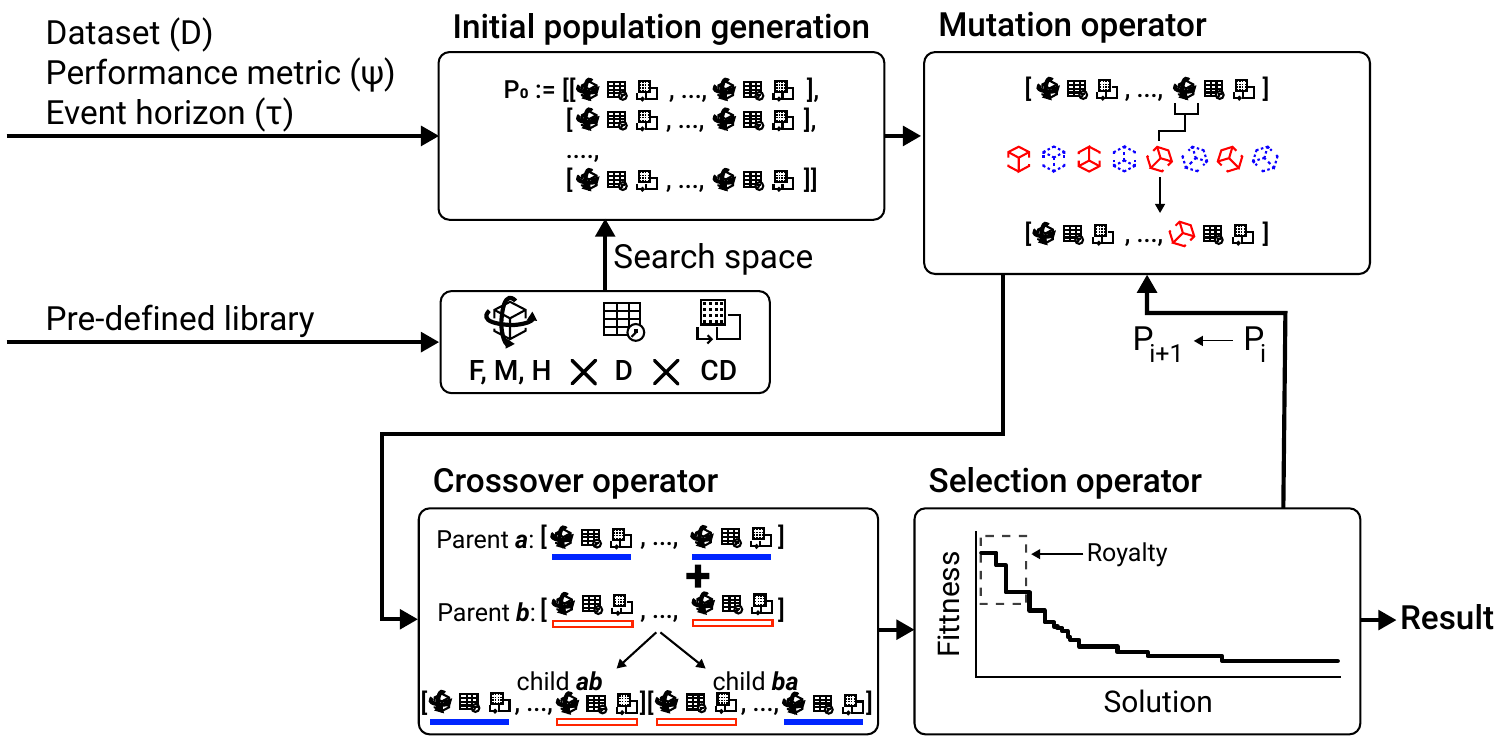}
    \caption{A schematic view of Algorithm 1. }
    \label{fig:algo}
\end{figure}

To this end, we proposed a genetic-based algorithm for finding a population of ML models to handle CD. The algorithm works as follows. First, a population of \(M\) solutions (i.e., a population of ML pipeline models populations) is generated at random such that each ML pipeline is chosen at random with a uniform distribution. Specifically, the ML pipelines as well as the CD detector algorithms are chosen at random with a uniform distribution. However, \(d_i\) are chosen by picking indexes \(t_1\) and \(t_2\) with a Poissonian distribution decaying from the latest sample in \(D(t)\) to the first one. In addition, the performance of the best solution from \(P_0\), as defined by Eq. \ref{eq:metric}, is computed. Now, for \(\psi \in \mathbb{N}\) generations, three operations are taking place, the \textit{mutation}, \textit{crossover}, and \textit{selection} operators to generate the next-generation population $P_{i+1}$. If \(M \in P_i\) is found to be better than the best performing \(M\) so far, it is becoming the best \(M\). The best-performing \(M\) during the entire process is returned as the answer of the model. The mutation operator is stochastically employed, for each \(M \in P_i\), with probability $\xi \in [0, 1]$. First, we randomly decide if to mutate the global ML model or one of the inner ML models in the population w.r.t. with probability $\zeta \in [0, 1]$. Either way, we replace one of the components of the ML pipeline model with another one, in a uniform manner distribution. For the case of \(d_i\), the start index (\(t_1\)) or the end index (\(t_2\)) is altered. First, the start or end index is randomly picked (with equal probability), and then a rounded value \(x\) which is distributed normally with \(\mu \in \mathbb{R}^+\) and \(\sigma \in \mathbb{R}^+\) as the mean and standard deviation of the distribution. Cross-over is employed for two \(M\) - $M_a$ and $M_b$ in population $P_i$ with the goal of creating two next-generation \(M_a\) and \(M_b\). We randomly choose a split-size $1 < s < min(|M_a|, |M_b|)$, and use it to split both ML populations, each to two random subsets - one of size $s$ and one of size $|M_a|-s$ and \(|M_b|-s\), respectively. i.e., $M_a = M_a^s \cup M_a^{|M_a|-s}$ and $M_b = M_b^s \cup M_b^{|M_b|-s}$. The cross-over then unifies complementing subsets from $a$ and $b$, creating $M_{ab}$ and $M_{ba}$ as follows \(M_{ab}=M_a^s \cup M_b^{|M_b|-s},~ M_{ba} = M_b^s \cup M_a^{|M_a|-s}\). The cross-over operation is performed over the entire population $P_i$. $P_i$ is first split into disjointed pairs of \(M\), and then the cross-over is performed on each such pair. Last, after employing mutation and cross-over, we employ the selection operator which forms the next-generation population $P_{i+1}$. We use the \textit{royalty tournament} operator~\cite{selection_operator}, which selects the best $\alpha \in [0, 1]$ ML populations from $P_i$ according to the fitness function $L(M, D)$. The rest of the ML populations are sampled (with repetitions) according to their fitness score, i.e., with probability \(p_{select}(M) = \frac{L(M, D)}{\sum_{M' \in P_i} L(M', D)}\). A pseudo-code representation of the proposed algorithm is presented in Algo \ref{algo:gen}. Fig. \ref{fig:algo} provides a schematic view of Algorithm 1. 

\begin{algorithm}[!ht]
	\caption{Genetic algorithm for population of ML models during concept drift} \label{algo:gen}
	\begin{algorithmic}[1]
	    \STATE $\text{\textbf{Input: }} \text{dataset }  (D), \text{performance measuring metric } (\psi), \text{event horizon } (\tau)$
	    \STATE $\text{\textbf{Output: }} \text{population of ML models }  (M)$
	    \STATE $P_1 \Leftarrow \text{generate } (n_t \sim [2, N] ) \text{ ML pipelines in random }$
	    \STATE $L_{best} \Leftarrow \forall M \in P_1: \max(L(M, D))$
	    \FOR{$\text{generation}~i \in [1, \dots, \phi]$}
  	        \STATE  $P_i \Leftarrow Mutation\_Operator(P_i, D. \psi, \tau)$
  	        \STATE  $P_i \Leftarrow Crossover\_Operator(P_i, D. \psi, \tau)$
  	        \STATE  $P_{i+1} \Leftarrow Selection\_Operator(P_i, D. \psi, \tau)$
  	        \IF{$\exists M \in P_i: \max(L(M, D)) > L_{best}$}
  	            \STATE $L_{best} \Leftarrow \max_M (L(M, D))$
  	        \ENDIF
  	    \ENDFOR
  \STATE return $argmax_{M \in P_i} \max(L(M, D)) $
	\end{algorithmic}
\end{algorithm}

\subsection{Improved version using the divide and conquer approach}
One can notice that the proposed model is a mixture of finding the optimal ML pipeline, \((f_i, m_i, h_i\) for each ML model in the population given a subset of the dataset, \(d_i\), finding the optimal subset of data and CD model, \((d_i, cd_i\), for each ML pipeline, and finding the global ML model as well as the CD model. Technically, the proposed model seeks to solve three dependent optimization tasks. 

Notably, the CD model for each ML pipeline in the ensemble does not have an effect on the rest of the components in the ML pipeline and only plays a role as an input for the global ML model. As such, once a subset of the dataset, \(d_i\) is terminated, the optimization task of each ML pipeline in the ensemble is independent. Moreover, once all  ML models in the ensemble are obtained, it defines a single optimization task to find the global ML pipeline. To this end, one can treat the finding of the CD models in the ensemble (\(\forall i: cd_i\)) as a constraint as part of the feature engineering component in the global model (\(f_g\)). Hence, the proposed model can be re-arranged as follows. First, divide the dataset, \(D(t)\), into subsets provided to the ML pipeline models in the ensemble. For each data subset (\(d_i\)), find the optimal ML pipeline model (\(f_i, m_i, h_i\)). Finally, uses these models, repeating the same process to find an optimal global ML pipeline model together with an adjustment CD model. 

In order to solve this representation of the model, the second and third tasks can be solved using an AutoML model designed especially for this task. For the data splitting task, we use Algo 1 with slight modification in which the algorithm assumes each model in the ensemble is defined only by its \(d_i\) component. 

\section{Experimental Setup}
\label{sec:experimental_setup}
Evaluating the proposed algorithm requires three components: dataset, baseline comparison, and performance evaluation metric. In this section, we outline these components which be used in the following section to assess the proposed algorithm. 

\subsection{Dataset curation}
\begin{figure}[!ht]
    \centering
    \includegraphics[width=0.99\textwidth]{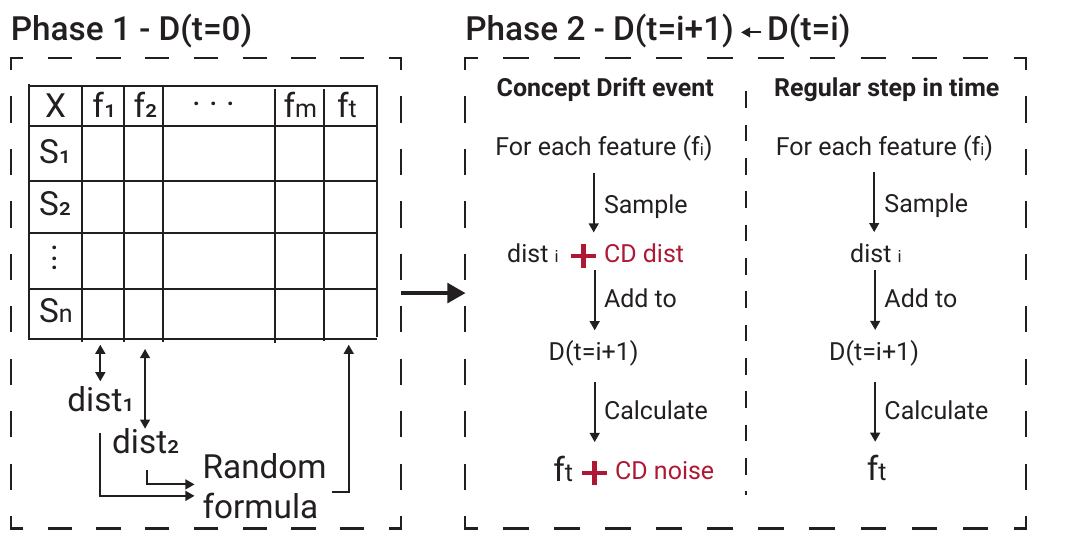}
    \caption{A schematic view of a dataset generation process for the experiments.}
    \label{fig:experiment_setup}
\end{figure}

In order to evaluate the proposed model, one is required to obtain a statistically large set of cases with CD in various levels and combinations. Moreover, as different CD can happen in multiple ways in parallel for different subsets (usually features) of the dataset, one should include this representation in the model. Due to the multiple moving parts and the challenge of detecting CD accurately \cite{cd_hard}, we used synthetic datasets for our analysis. Intuitively, when developing an ML model some dataset is already established based on data gathered, marked by \(D(0)\). This can be considered the initial (or first) phase of the data curation process. At this point, we can assume no CD is present and there is some connection between the target and source features. After this point, as part of the second phase, more data streams to the dataset over time with discrete steps. At each step in time, there is either a CD event or not. If no CD is present, more data, according to some distribution (which is unknown for the ML model) is generated at random and added to the dataset, \(D(t)\), at time \(t\). Otherwise, a concept drift event is associated with the change of a subset (or all) features' distributions of the dataset during a period of \(\Delta t \in \mathbb{N}\). In addition, CD can also introduce change to the connection between the source and target feature which for the ML reflected like noise. Fig \ref{fig:experiment_setup} shows a schematic view of the dataset generation process for the experiments. 

Formally, a dataset (\(D(t)\)) is generated as follows. First, the initial data, \(D(0)\), is generated followed by a function \(F: D(t) \rightarrow D(t+1)\) which accepts the dataset at a point in time, \(t\), and returns the same dataset for the next point in time, \(t+1\). For the initial data, a random number of samples, \(s \in \mathbb{N}\), and features, \(f \in \mathbb{N}\) is picked at random from some pre-defined distribution. Then, for each feature, a random distribution is chosen, from a pre-defined set of known distributions, and \(s\) samples are obtained from it. In order to obtain a meaningful regression task, the target feature of each dataset is obtained by defining a random function (contracted from a combination of polynomials, exponential, trigonometric, logarithmic, and step-wise functions) which gets as an input the other features of the dataset. To be exact, the function is constructed by randomly picking a topology for an expression tree where each leaf node is a feature from the dataset and each decision node is a function from a pre-defined set of functions \cite{scimed}. For the function, \(F\), a sequence of CD events is set alongside a default behavior. For the default behavior, which is used when there are no CD events for a specific step in time, a random number \(\zeta \in \mathbb{N}\) of new samples is generated for each feature using the current distributions associated with each feature and then the target feature is computed for these samples using the current formula. In a complementary manner, a CD event at some time changes the distribution of a subset (or all) features in the dataset over a period of \(\Delta t\) steps in time as well as changing the formula used to calculate the target feature. 

In order to obtain a robust representation of an algorithm's performance, we generated multiple datasets for each explored configuration. The hyperparameters used as part of the data generation process are summarized in Table \ref{table:data_generation_hyperparameters}. 

\begin{table}[!ht]
\centering
\begin{tabular}{p{0.15\textwidth}p{0.5\textwidth}p{0.35\textwidth}}
\hline \hline
\textbf{Hyperparameter} & \textbf{Description} & \textbf{Value range} \\ \hline \hline
  \(n, m\)  & The number of samples and features in the initial dataset & \([10^2, 10^5, 3-50]\) \\
  \(dist_i\) & The distribution of a feature in the dataset & Normal, Exponential, Binomial, Geometric, Benford, Uniform   \\
  \(\eta\) & Formula topology's size & \([n, 10n]\) \\ 
  \(\tau\) & Formula functions & \([+, -, *, exp, log, inv, sin, scalar, step-wise]\) \\ \hline \hline
\end{tabular}
\caption{The hyperparameters used as part of the data generation with their value ranges.}
\label{table:data_generation_hyperparameters}
\end{table}

For the experiments, we used the ML components available in the Scikit-learn library \cite{scikit_learn}. In addition, we used the CD algorithms provided by the Frouros library \cite{cd_library}. 

\subsection{Baseline algorithms}
For the comparison of the proposed algorithm with other configurations, we establish three baseline models. First, a \textit{single-random} model where only a single ML pipeline and CD model are chosen at random. Second, a \textit{multi-random} model where the proposed ML population with global ML model is used but each of them is picked at random. Third, a single ML pipeline with a CD algorithm which is obtained using TPOT and brute force, respectively. In addition, the proposed model and its improvement are considered in the evaluation as the fourth and fifth candidates, respectively. 

\subsection{Performance metric}
Assessing the performance of the overall model requires measuring the performance of the model on various tasks over time. To this end, for a specific dataset, \(D(t)\), and a time frame of interest, \(t_0-t_1\) (\(t_1 > t_0\)), the performance of the model is defined to be the average performance over all steps in time. Formally, the ML pipeline with CD adoption model's performance is defined to be \cite{metric}:
\begin{equation}
    \theta(M) := \frac{1}{(t_1 - t_0)N} \sum_{i=0}^{N} \sum_{j = t_0}^{t_1} \psi \big ( M, D_j(i) \big )   \label{eq:ml_cd_performance}
\end{equation}

\section{Results}
\label{sec:results}
In this section, we outline the results of the experiments. First, we compare the proposed model to other single- and multi- ML pipeline solutions. Afterward, we explore the robustness and sensitivity of the best solution to the concept drift rate, dataset's size, and dataset's complexity. 

\subsection{Performance comparison}
\label{sec:performance_comparison}
Table \ref{table:comparision} presents the comparison between the five models with the metric presented in Eq. (\ref{eq:metric}) for four cases - shift, moving, mixed, and random CD. For the shift concept drift, the datasets are generated with a single CD starting at a random point in time and have a random drift rate between \(0.1\) and \(0.2\), chosen in a uniformly distributed manner. Similarly, the moving CD case is identical to the shift case but with a drift rate between \(0.01\) and \(0.02\). For the mixed case, a random number of CD can occur ranging between 2 and 10 with the constraint that at least one would be a shift CD and another is a moving CD. Finally, the random case is like the mixed case but without any constrain on the CD type. For each case, we used \(n=1000\) datasets. 

\begin{table}[!ht]
\centering
\begin{tabular}{lllll}
\hline \hline
\textbf{Model} & \textbf{Shift} & \textbf{Moving} & \textbf{Mixed} & \textbf{Random} \\ \hline \hline
Single-random      & 0.59 (0.13) & 0.65 (0.11) & 0.47 (0.18) & 0.48 (0.18) \\
Multi-random       & 0.63 (0.10) & 0.69 (0.08) & 0.56 (0.15) & 0.58 (0.16) \\
TPOT               & 0.68 (0.09) & 0.75 (0.06) & 0.52 (0.16) & 0.51 (0.16) \\
Proposed           & 0.74 (0.07) & 0.80 (0.06) & 0.64 (0.11) & 0.64 (0.11) \\
Proposed improved  & 0.76 (0.06) & 0.81 (0.05) & 0.67 (0.10) & 0.67 (0.10) \\ \hline \hline
\end{tabular}
\caption{Comparison of different ML pipelines with CD detection models for four different CD cases - shift, moving, mixed, and random. The results are shown as the mean with the standard deviation in brackets of Eq. (\ref{eq:metric}). }
\label{table:comparision}
\end{table}

\subsection{Sensitivity analysis}
In order to evaluate the proposed model's performance in different settings, we explore the performance of both the proposed model and its improved version on different concept drift rates, dataset sizes, and dataset's complexity. For the dataset's complexity, we adopted the metric proposed by \cite{assaf} which associated the dataset's complexity with its non-linearity measured by by using the \(1-R^2\) value obtained from a linear regression model trained on the dataset.

\subsubsection{Drift rate}
Table \ref{table:sens_drift_rate} summarizes the performance, in terms of Eq. (\ref{eq:metric}), as the mean of \(n=1000\) cases for each drift rate. One can notice an average decrease in the performance of the model as the drift rate increases. Nonetheless, for a drift rate of 0.2, the performance of both models slightly increases as the shift CD is clearer and easier to detect by the CD algorithms in the ensemble. Importantly, we allow between 2 and 10 CDs for each sample such that all of them have the same drift rate.

\begin{table}[!ht]
\centering
\begin{tabular}{lcccccccc}
\hline \hline
\textbf{Model \textbackslash Drift rate} & 0.01 & 0.025 & 0.05 & 0.075 & 0.1 & 0.15 & 0.2 \\ \hline \hline
Proposed & 0.81 (0.06) & 0.80 (0.06) & 0.79 (0.06) & 0.77 (0.07) & 0.73 (0.07) & 0.74 (0.07) & 0.75 (0.06) \\
Proposed improved & 0.81 (0.05) & 0.81 (0.05) & 0.80 (0.06) & 0.77 (0.06) & 0.75 (0.06) & 0.76 (0.06)  & 0.77 (0.06) \\ \hline \hline
\end{tabular}
\caption{Sensitivity analysis for the proposed model and its improved version in terms of drift rate.}
\label{table:sens_drift_rate}
\end{table}

\subsubsection{Dataset size}
Table \ref{table:sens_complexity} summarizes the performance as the mean of \(n=1000\) cases for datasets with different initial sizes and growth rates. The growth rate is the number of new samples added to the dataset in each step in time. For this analysis, we used the random CD case (see Section \ref{sec:performance_comparison}). The analysis shows that for datasets with relatively large initial sizes (\(>10^4\)), the performance over the growth rate is more stable for both models. However, for relatively small initial sizes (\(<10^3\)), the growth rate has the dominant effect on the model's performance. Overall, a tendency for more data is more beneficial for the proposed model. 

\begin{table}[!ht]
\centering
\begin{tabular}{lc||cccc}
\hline \hline
\textbf{Model} & \textbf{Initial size \textbackslash growth rate} & \(10^0\) & \(10^1\) & \(10^2\) & \(10^3\) \\ \hline \hline
\multirow{4}{*}{Proposed} & \(10^2\) & 0.42 (0.18) & 0.44 (0.16) & 0.50 (0.14) & 0.57 (0.13) \\
                         & \(10^3\)  & 0.49 (0.17) & 0.49 (0.15) & 0.53 (0.14) & 0.57 (0.13) \\
                         & \(10^4\)  & 0.62 (0.12) & 0.62 (0.12) & 0.62 (0.12) & 0.63 (0.11) \\
                         & \(10^5\)  & 0.64 (0.11) & 0.64 (0.11)  & 0.64 (0.11)  &  0.64 (0.11) \\ \hline 
\multirow{4}{*}{Proposed improved} & \(10^2\)   & 0.44 (0.19) & 0.45 (0.18) & 0.49 (0.17) & 0.59 (0.12) \\
                                     & \(10^3\) & 0.49 (0.17) & 0.49 (0.17) & 0.50 (0.16) & 0.59 (0.12) \\
                                     & \(10^4\) & 0.65 (0.12) & 0.65 (0.12) & 0.65 (0.11) & 0.67 (0.10) \\
                                     & \(10^5\) & 0.67 (0.10) &  0.67 (0.10)  &  0.67 (0.10) &  0.67 (0.10) \\  \hline \hline
\end{tabular}
\caption{Sensitivity analysis for the proposed model and its improved version for different initial size and growth rates. }
\label{table:sens_sizes}
\end{table}

\subsubsection{Dataset's complexity}
Table \ref{table:sens_complexity} summarizes the performance as the mean of \(n=1000\) cases for datasets with different complexity levels. For this analysis, we used the random CD case (see Section \ref{sec:performance_comparison}). This analysis shows that while more \say{complex} dataset has lower results in absolute terms due to the poorer performance of ML in general, the proposed model is only slightly negatively affected by the dataset's complexity. 

\begin{table}[!ht]
\centering
\begin{tabular}{llccccc}
\hline \hline
\textbf{Model} & \textbf{Comparison} & \textbf{0.1} & \textbf{0.3} & \textbf{0.5} & \textbf{0.7} & \textbf{0.9} \\ \hline \hline
\multirow{2}{*}{Proposed} & Absolute & 0.71 (0.11) & 0.78 (0.07) & 0.81 (0.06) & 0.80 (0.05) & 0.79 (0.06) \\ 
 & Relative & 0.89 (0.04) & 0.88 (0.04) & 0.90 (0.03) & 0.89 (0.04) & 0.90 (0.04) \\ \hline
\multirow{2}{*}{Proposed improved} & Absolute & 0.73 (0.09) & 0.81 (0.05) & 0.84 (0.05) & 0.82 (0.06) & 0.81 (0.05) \\
 & Relative & 0.92 (0.03) & 0.90 (0.04) & 0.90 (0.04) & 0.90 (0.04) & 0.91 (0.04) \\ \hline \hline
\end{tabular}
\caption{Sensitivity analysis for the proposed model and its improved version in terms of the dataset's complexity. The \textit{absolute} comparison does not account for the ML pipeline performance in the substance of CD while the \textit{relative} divides the result of Eq. (\ref{eq:metric}) by \(\psi(M, D(0))\).}
\label{table:sens_complexity}
\end{table}

\subsubsection{AutoML library usage}
Table \ref{table:sens_automl} summarizes the performance as the mean of \(n=1000\) cases where the improved proposed model is utilized with different AutoML libraries. For this analysis, we used the random CD case (see Section \ref{sec:performance_comparison}). One can notice that all four examined libraries produced similar results with TPOT and AutoGluon being the best and worse, respectively, on average.

\begin{table}[!ht]
\centering
\begin{tabular}{lcccc}
\hline \hline
\textbf{Model \textbackslash AutoML library} & TPOT & AutoSklearn & AutoGluon & PyCaret \\ \hline 
Proposed improved & 0.67 (0.10) & 0.65 (0.09) & 0.64 (0.11) & 0.64 (0.09) \\\hline \hline
\end{tabular}
\caption{Sensitivity analysis for the proposed model and its improved version in terms of drift rate.}
\label{table:sens_automl}
\end{table}

\section{Discussion}
\label{sec:discussion}
In this study, we investigated the usage of ensemble ML models to handle CD in datasets that increase over time. In particular, we proposed a novel instance of the GA approach for an ML pipeline with CD detection algorithm ensemble. Using this structure, the overall model is more robust for different CD events. Moreover, we show that one can further improve the proposed algorithm by integrating existing off-the-shelf automatic ML approaches. Overall, the proposed model allows to use of existing ML and CD algorithms to bolster the resilience of ML models in the face of CD in realistic scenarios. 

To be exact, the two-level ML model proposed in this study holds promise in addressing the dynamic challenges posed by CD by introducing a global ML model with a CD detector operating as an ensemble model for a population of ML pipeline models which also can be adopted by an adjusted CD detection algorithm. Hence, this study takes a novel approach compared to the one-model-to-rule-them-all approach currently governing the field \cite{cd_discussion_1,cd_discussion_2}. Simply put, the ensemble structure utilized by the proposed model allows individual ML to autonomously adapt to subset-specific changes while feeding the prediction of each model and how well it is operating, as indicated by its adjusted CD detector, showcasing the adaptability of GAs in governing diverse data subsets.

Indeed, Table \ref{table:comparision} shows that the proposed solution provides a more robust solution, on average, for a large number of datasets compared to a single ML pipeline with a CD algorithm obtained using an AutoML library (such as TPOT) and brute-force of multiple (12) CD algorithms. For the most difficult and realistic settings where the number, as well as nature or the CD, are unknown (i.e., the \textit{random} case) the proposed model improves the performance performance of the ML model by up to 0.13 while also reducing the diversity in the results between datasets from 0.16 to 0.11, indicating a more robust and consistent results across different datasets.  

Moreover, we explore the performance of the proposed model over different settings. First, Table \ref{table:sens_drift_rate} shows that the proposed model performs better from moving CD rather than shifting CD. This outcome aligns with the behavior of other solutions \cite{dis_sens_1,dis_sens_2}. Second, Table \ref{table:sens_sizes} reveals that the proposed model performs worse on small-size datasets, which is also a well-known phenomenon in the realm of ML \cite{dis_sens_3,dis_sens_4}. However, once enough data is available, the proposed model performs similarly. To this end, Table \ref{table:sens_complexity} continues the same line where more \say{complex} datasets resulted in worse performance in absolute terms as the underline ML pipeline models are performing worse. Nonetheless, in relative terms, the complexity of the dataset does not play much role in the context of handling CD. Finally, Table \ref{table:sens_automl} shows that the improved version of the proposed model is only slightly affected by the autoML library used, at least from the popular autoML libraries currently available, and one can choose its preferred autoML library. 

This research is not without limitations. First, the proposed solution is evaluated on synthetic data due to the challenge and resources required to find real-world cases of CD in large numbers and for a wide range of CD behaviors. Hence, the proposed results should be taken with caution and future work should re-evaluate the proposed method using a large number of real-world CD datasets. Second, the proposed method provides each ML model in the population a subset of the dataset, \(D(t)\), which is continuous, ignoring the more generic cases where one can union several continuous subsets of \(D(t)\) which can improve the performance the ML model alone and the entire performance, as a whole. One can investigate the contribution of such an extension to the proposed method's performance. Third, GA in general, and for the discussed case, in particular, often requires tuning several hyperparameters, such as population size, crossover rate, and mutation rate to achieve their optimal performance \cite{ga_tune_1,ga_tune_2}. In this study, we chose such hyperparameter values through a manual trial and error approach which probably do not result in the optimal values. Future studies may explore the sensitivity of the proposed GA-based approach to variations in these parameters and find the optimal hyperparameter values with respect to assumptions on the CD dynamics or the data itself. Fourth, the presented experiments focused on regression tasks, ignoring classification tasks. Extending the evaluation to classification tasks can shed more light on the performance of the proposed model in a wider context. Finally, following a more general trend \cite{k_1,k_2,k_3,k_4,k_5}, one can reduce the search space of the proposed task and therefore improve the proposed algorithm by introducing knowledge about which ML models are appropriate to which types of CD and the best matching between an ML model and a CD detector. 

This study marks a significant stride in fortifying ML models against the persistent challenges posed by CD through the approach of a two-level ensemble of ML models working together with the CD application of genetic algorithms GAs. The adaptability demonstrated, particularly in discrete spaces where traditional optimization methods face limitations, highlights the promising role of GAs in addressing the nuanced demands of evolving data distributions. While the research outcomes are encouraging, avenues for refinement and future exploration are recognized. Extending the generalizability across diverse domains, conducting a more comprehensive sensitivity analysis, and optimizing computational complexity for broader applicability should be key considerations. In essence, this research contributes to the ongoing discourse on adaptive learning systems, leveraging GAs in an ensemble ML context to navigate the challenges presented by dynamic data landscapes. 

\section*{Declarations}
\subsection*{Funding}
This research did not receive any specific grant from funding agencies in the public, commercial, or not-for-profit sectors. 

\subsection*{Conflicts of interest/Competing interests}
None. 

\subsection*{Materials availability}
The materials used for this study are available from the author upon reasonable request. 

\bibliography{biblio}
\bibliographystyle{unsrt}

\end{document}